\pdfoutput=1

\documentclass[11pt]{article}

\usepackage{emnlp2021}

\usepackage{times}
\usepackage{latexsym}
\usepackage{lipsum}
\usepackage{cuted, caption}
\usepackage{graphicx}

\usepackage{amsmath,amsfonts,amssymb,mathtools,amsthm}
\usepackage{stmaryrd}
\usepackage{hyperref}
\usepackage{array,multirow}

\usepackage[T1]{fontenc}

\usepackage[utf8]{inputenc}

\usepackage{microtype}

\title{Do Language Models Know the Way to Rome?}

\author{Bastien Liétard \\
  University of Lille \\
  \texttt{bastien.lietard.etu} \\ \texttt{@univ-lille.fr} \\\And
  Mostafa Abdou and Anders Søgaard  \\
  University of Copenhagen \\
  \texttt{\{madbou,soegaard\}@di.ku.dk}}
  
\definecolor{addg}{rgb}{0.100, 0.700, 0.100}
\definecolor{modb}{rgb}{0.100, 0.100, 0.700}

\begin{document}
\maketitle
\begin{abstract}
The global geometry of language models is important for a range of applications, but 
language model probes tend to evaluate rather local relations, for which ground truths are easily obtained. In this paper we exploit the fact that in geography, ground truths are available beyond local relations. In a series of experiments, we evaluate the extent to which language model representations of city and country names are isomorphic to real-world geography, e.g., if you tell a language model where Paris and Berlin are, does it know the way to Rome? We find that language models generally encode limited geographic information, but with larger models performing the best, suggesting that geographic knowledge {\em can} be induced from higher-order co-occurrence statistics. 
\end{abstract}

\section{Introduction}
Language models (LMs) are fundamental building blocks in state-of-the-art natural language processing. These models are trained to predict words in context, and as a side product, they learn to compress higher-order co-occurrence statistics to represent the distributional properties of words and phrases. 

It is well-documented that the representations of modern language models encode some syntactic \cite{tenney-etal-2019-bert} and semantic knowledge (\cite{NEURIPS2019_159c1ffe}), as well as some real-world knowledge \cite{davison-etal-2019-commonsense,petroni-etal-2019-language, jiang2020can, roberts2020much}, i.e., some knowledge base relations can be extracted directly from language models. Much of this information was available in older language models such as Word2Vec, too \cite{mikolov2013}. 

The probes that have been designed for the above studies, however, only probe for one-dimensional information (e.g., part of speech or semantic class) or very local relations (e.g., is Daniel Kehlmann a writer?). It is an open question whether higher-order co-occurrence statistics found in textual corpora can induce representations show isomorphism to human mental representations of entities and relations. To test this we operationalize a dataset of geographic information. Geographic knowledge is a domain in which we can go beyond local relations, since we can access global representations of geographic knowledge (to the extent it is isomorphic to the physical world).  

Cross-lingual language model alignment provides additional motivation for probing for geographic knowledge: We know that language models of different languages are sometimes near-isomorphic  \cite{soegaard-etal-2018,vulic-etal-2020-good}, and it is tempting to think that this is because language models reflect real-world structures, and that the (occasional) near-isomorphism of language models is simply the result of the fact that we use language(s) to talk about the same world. Is the isomorphism of language models a result of them being in part isomorphic to the physical world? 

Why would language models potentially encode physical geometry? Higher-order co-occurrence statistics can be surprisingly informative. Copenhagen and Malm{\"o} are in different countries, but connected by a bridge, often referred to as {\em the bridge between Copenhagen and Malm{\"o}}. Other cities belong to the same municipalities and their names therefore co-occur with the name of the municipality. Other cities were, for instance, potentially impacted by the same natural disaster or part of the same development project. By distilling thousands, or maybe hundreds of thousands, of such co-occurrences, we conjecture that language models might be able to induce somewhat fine-grained maps of physical geometry. 

In this work, we study the extent to which geographical information is encoded by language models. Based on geographic information from multiple sources, we train probing models on language model representations and evaluate their ability to predict the locations of cities and countries, as well as their ability to predict (one-dimensional) population numbers and (local) neighbor relations, relative to control probe performance levels. 

\paragraph{Contributions} We collect geographic information from multiple sources and design experimental protocols to probe language models for such information. We evaluate two BERT models \cite{devlin2019} of different sizes, two RoBERTa models \cite{liu2019} of different sizes, as well as GPT-2 \cite{radford2019} and Word2Vec \cite{mikolov2013} across three different tasks and various protocols. Our results show that modern language models do not encode much more geographic information than Word2Vec, but that larger (non-autoregressive) language models encode more information than their smaller counterparts, suggesting that such information is nevertheless available through higher-order co-occurrence statistics. 

\section{Related Work}
Word representations based on distributional statistics have been theorised to capture a wide range of information \cite{schutze1992dimensions}. To evaluate this, a considerable body of literature has made use of semantic similarity, relatedness, and analogy datasets \cite{agirre2009study, bruni2012distributional, baroni2014don, faruqui2014retrofitting, hill2015simlex, drozd2016word, abdou2018mgad}. Asking a broader question, \citet{rubinstein2015well} investigated the types of semantic information which are encoded by different classes of word embedding models, finding that taxonomic properties (such as animacy) are well-modelled. In a similar direction, \citet{collell2016image} and \citet{lucy2017distributional} draw on semantic norm datasets to test how well these models can encode a range of perceptual and conceptual features.

In the context of the neural language models, several recent works such as  \citet{davison-etal-2019-commonsense} and \citet{petroni-etal-2019-language} have attempted to extract factual and commonsense knowledge from them by posing knowledge base triplets as close statements which are used to query the models. Most related to this work, \citet{forbes2019neural}  investigate whether LMs can learn physical commonsense through language. They find that LM representations do indeed encode information regarding object properties (e.g., \textit{bananas are yellow}), and affordances (e.g. \textit{bananas can be eaten}) but they do not to capture the more subtle interplay between the two. 

Studies investigating the geometry of word representations have focused on intrinsic dimensionality and subspaces \cite{yaghoobzadeh2016intrinsic, coenen2019visualizing}, embedding concentration in narrow cones and anisotropy \cite{mimno2017strange, Ethayarajh2019}, and comparisons to the geometry of cognitive measurements or perceptual spaces \cite{abnar2019blackbox, abdou2021language}.  In this work we investigate the degree of isomorphism between language model representations of geographic location names and their real-world counterparts. See \citet{vulic-etal-2020-good} for a general discussion of isomorphism across language models, what explains this, and what it depends on. 


\section{Methodology}

We probe language model representations of city and country names. 
Below, we propose three classification/regression probing tasks, as an well as an analysis based on (relational) similarity. 

\subsection{Probing tasks}

\paragraph{Task 1: Predicting geo-coordinates}
We probe LMs' representation about whether they convey information about the actual position of the location. To this extent, we train a model to predict GPS coordinates (latitude and longitude) given the location's representation through language models. We evaluate prediction error by computing the distance in kilometers between the predicted GPS point and the expected point, and compute the average of error distances. Since we use linear models (Lasso\footnote{We also include an MLP regressor for comparison.}) to predict geo-coordinates, we effectively evaluate the isomorphism of language model representations to the physical world. Our target variable is two-dimensional, and we evaluate non-local relations.  

\paragraph{Task 2: Predicting population sizes}
In the physical world, cities are characterized not only by their location, but also by their size and population. There is indeed a major difference between {Shanghai (27M people)} and {Worcester (101K people)}. Given the representation of a country name or a city name, we train a probe to predict the population living in the location. We evaluate the performance by computing the \textit{mean squared error}. This is a one-dimensional attribute included here for comparison. 

\paragraph{Task 3: Predicting neighboring countries}
Borders between countries are also an important geographic relation that is closely related to a country's international policy. It also provides information about the continental situation ; indeed, a country with no land borders must be an island. To probe language models for neighboring relations, we train an classifier to predict whether two countries share a border or not, given the pair of representations. We report the probe accuracy. Neighboring relations are local and included here for comparison.

\subsection{Control tasks and scores}

\citet{hewitt2019} have shown the importance of a control task to balance the probe accuracy. We construct our control tasks by randomly permuting the target variables and train the model on this randomly-permuted dataset. We repeat this control task 10 times and take the mean error/accuracy.

\paragraph{Probe classifier selectivity}
For the country borders classification task, we define the \textit{probe selectivity} as the difference between the probe accuracy on the original task and the accuracy on the control task \cite{hewitt2019}.

\paragraph{Probe error reduction}
Similarly, we define the \textit{probe error reduction}, which measures how well the probe performs on the original task compared to the control task, by comparing positive error measures of both tasks through error reduction, i.e., the proportion of control task error that is reduced in our probe.

$$
PER = 1 - \frac{\textit{Probing task error}}{\textit{Control task error}}
$$

Less formally, if the PER is between 0 and 1, its upper bound, it indicates to what extent the probe is doing better on the original task than on the control task ; a negative PER indicates the probe is doing worse on the original task than on the control task. We use PER for regression tasks rather than absolute error or MSE, because figures are more easily interpretable when properly baselined. Note that if absolute error is 0, PER is 1; if absolute error is as high as the random baseline, PER is 0. Under the assumption performance is better than random, PER thus ranges between 0 and 1. 

\subsection{Similarity Analysis}
For each language models, we compute the cosine similarity between each pair of cities. Our hypothesis is that cities belonging to the same country will be more similar than two cities from different countries. We report both average similarities (cities in the same country and cities in different countries) and also compute the histogram of the distribution of these two sets of similarities.

\section{Experimental Settings and Data}

\paragraph{Language models} We compare three different state-of-the-art contextualized LMs, namely BERT \citep{devlin2019}, RoBERTa \citep{liu2019} and GPT-2 \citep{radford2019}. A major difference between GPT-2 and the two other models is that it is trained to be unidirectional (\emph{i.e.} using next-token language modelling), whereas BERT and RoBERTa are trained to be bidirectionnal (\emph{i.e.} using masked language modelling). For BERT and RoBERTa, we used both \texttt{base} ($L=12$ layers and $D=768$ output dimensions) and \texttt{large} ($L=24$ layers and $D=1024$ output dimensions) version, so we can study the influence of the model's size. We probe widely-used pre-trained versions of these models, as they have been trained in their respective original papers. Both versions of BERT and RoBERTa have been pre-trained on the BookCorpus (800M words) and English Wikipedia (2.5B words). GPT-2 is also trained on the Book Corpus.

\paragraph{Geographic data} We get a list of names of cities around the world having a population of 100K or more people, with their GPS coordinates (Latitude and Longitude), their country and their population, from a MaxMind database\footnote{\url{https://www.kaggle.com/max-mind/world-cities-database}}. Country geo-coordinates correspond to the centroid of each country. We obtain population size data from the  United Nations' Human Development Data Center\footnote{\url{http://hdr.undp.org/en/data}}, given in millions of inhabitants. Our datasets are then composed of 3527 cities and 249 countries and territories.
\vspace{7pt}

\paragraph{Representations extraction} Since the LMs provide \textit{contextualised} token representations, we provide three linguistically-basic contexts : \textit{He lives in} $X$, \textit{She moved to} $X$ and \textit{I come from} $X$, replacing $X$ with either a city name or a country name. For each context, we extract the representation corresponding to the name by averaging the hidden states of the last 4 layers of the LMs. Following \citet{bommasani2020}, we use a mean pooling of subwords' representations when necessary.
\vspace{7pt}

\paragraph{Word2Vec} To compare static and contextualized models, we also use pre-trained Word2Vec \textit{static} representations learned over the Google News dataset (150B words), provided by the \texttt{gensim}\footnote{\url{https://radimrehurek.com/gensim/}} library. The representational vectors are in 300 dimensions.
\vspace{7pt}

\begin{table*}[t]
    \centering
    \begin{tabular}{c|cc||c|cc|c|cc}
        Task & Probe & Dataset & Word2Vec & BERT & BERT-L & GPT-2 & RoBERTa & RoBERTa-L \\\hline\hline
        \parbox[t]{0.4cm}{\multirow{4}{*}{\rotatebox[origin=c]{90}{GPS}}} & \multirow{2}{*}{MLP} & city & \textbf{0.666} & 0.479 & 0.585 & 0.424 & 0.466 & 0.517 \\
        & & country & 0.57 & 0.424 & \textbf{0.58} & 0.422 & 0.405 & 0.468\\\cline{2-9}
        & \multirow{2}{*}{Lasso} & city & 0.498 & 0.454 & \textbf{0.555} & 0.364 & 0.398 & 0.466\\
        & & country & 0.395 & 0.394 & \textbf{0.465} & 0.349 & 0.378 & 0.454 \\\hline\hline
        \parbox[t]{4mm}{\multirow{4}{*}{\rotatebox[origin=c]{90}{Population}}} & \multirow{2}{*}{MLP} & city & 0.507 & 0.694 & 0.704 & 0.624 & 0.696 & \textbf{0.707}\\
        & & country & 0.618 & 0.59 & \textbf{0.63} & 0.568 & 0.528 & 0.505\\\cline{2-9}
        & \multirow{2}{*}{Lasso} & city & \textbf{0.628} & 0.539 & 0.55 & 0.447 & 0.403 & 0.531\\
        & & country & -0.242 & 0.116 & 0.276 & \textbf{0.372} & 0.078 & 0.116\\\hline

    \end{tabular}
    \caption{Probe error reduction. ``BERT-L'' and ``RoBERTa-L'' denote \texttt{large} versions of BERT and RoBERTa. Column name ``Probe'' designates the different regression architectures used for probing. For GPS coordinates, error reduction is computed with average error in kilometers, whereas it is computed using MSE for Population. Control tasks are performed on $n=10$ trials, and the control error is obtained by averaging all trials errors.}
    \label{tab:control_scores}
\end{table*}

\paragraph{Probe models} For the classification problem (task 3), we use a 100-units single-layered MLP as a classifier. As a probe for regression task (1 and 2), we train both a Lasso regressor with a L2-penalty of $\alpha$ and a single hidden layer MLP with 100 units. We arbitrarily choose $\alpha=1$, except for task 1 for which we remarked that $\alpha=0.5$ significantly decreased the training time without affecting to much the error ratio. No other hyperparameters-tuning was done, since we are not interested in designing the best task-specific models. Probes are trained on a random split of 80\% of the dataset, and evaluate on the 20\% remaining.

\begin{table}[t]
    \centering
    \begin{tabular}{c|ccc}\hline
        & \multicolumn{2}{c}{Accuracy} & Probe \\
        Model & Prb. & Ctrl. & Selectivity\\\hline\hline
        Word2Vec & 0.849 & 0.49 & 0.36\\BERT & 0.856 & 0.51 & 0.34\\BERT-L & \textbf{0.873} & 0.51 & \textbf{0.37}\\GPT-2 & 0.808 & 0.51 & 0.3\\RoBERTa & 0.817 & 0.51 & 0.31\\RoBERTa-L & 0.843 & 0.52 & 0.32\\\hline
    \end{tabular}
    \caption{Probe accuracy (Prb.), control accuracy (Ctrl.) and probe selectivity for country borders prediction.}
    \label{tab:border_probe}
\end{table}

\section{Results}

In Table \ref{tab:control_scores}, probe error reduction scores are displayed for both regression tasks, namely Task 1 and 2. The full list of performances (error distances in kilometers) for Task 1 is provided in the appendix \ref{sec:task1perf}. In Appendix \ref{sec:task2perf}, error values for population prediction with countries are reported. Since the number of countries is relatively small, we perform a \textit{5-fold cross-validation} with this dataset. 

When predicting a location's GPS coordinates, all models have an error reduction score significantly larger than 0, across both families of regressors and across cities and countries. The Word2Vec model especially has often the lowest error and a high PER, showing no real weakness compared to contextualised models. GPT-2 has the lowest PER and the worst error value, across almost all tasks and probes. The two regressors globally lead to the same error values, even though MLP allows a slightly bigger error reduction than Lasso. We can also note that, for both BERT and RoBERTa, increasing model size (moving from \texttt{base} models to \texttt{large}), one increases the gap between the performance on the task and on the random control, reducing overall error. Finally, we can also see that the models are generally not very accurate for predicting GPS coordinates, leading to an average error of 2'500-5'000 kilometers. It is relatively accurate continent-wise, but not precise. 

In sum, we can conclude that it is non-trivial to learn linear maps from language models to physical world geo-coordinates, presumably because representations are non-isomorphic. On the other hand, the improvements from larger models seem to suggest that geographical knowledge is available from higher-order co-occurrence statistics. Static model also provide a strong baseline compared to contextualised models.

\begin{figure*}[t]
    \centering
    \includegraphics[width=8cm]{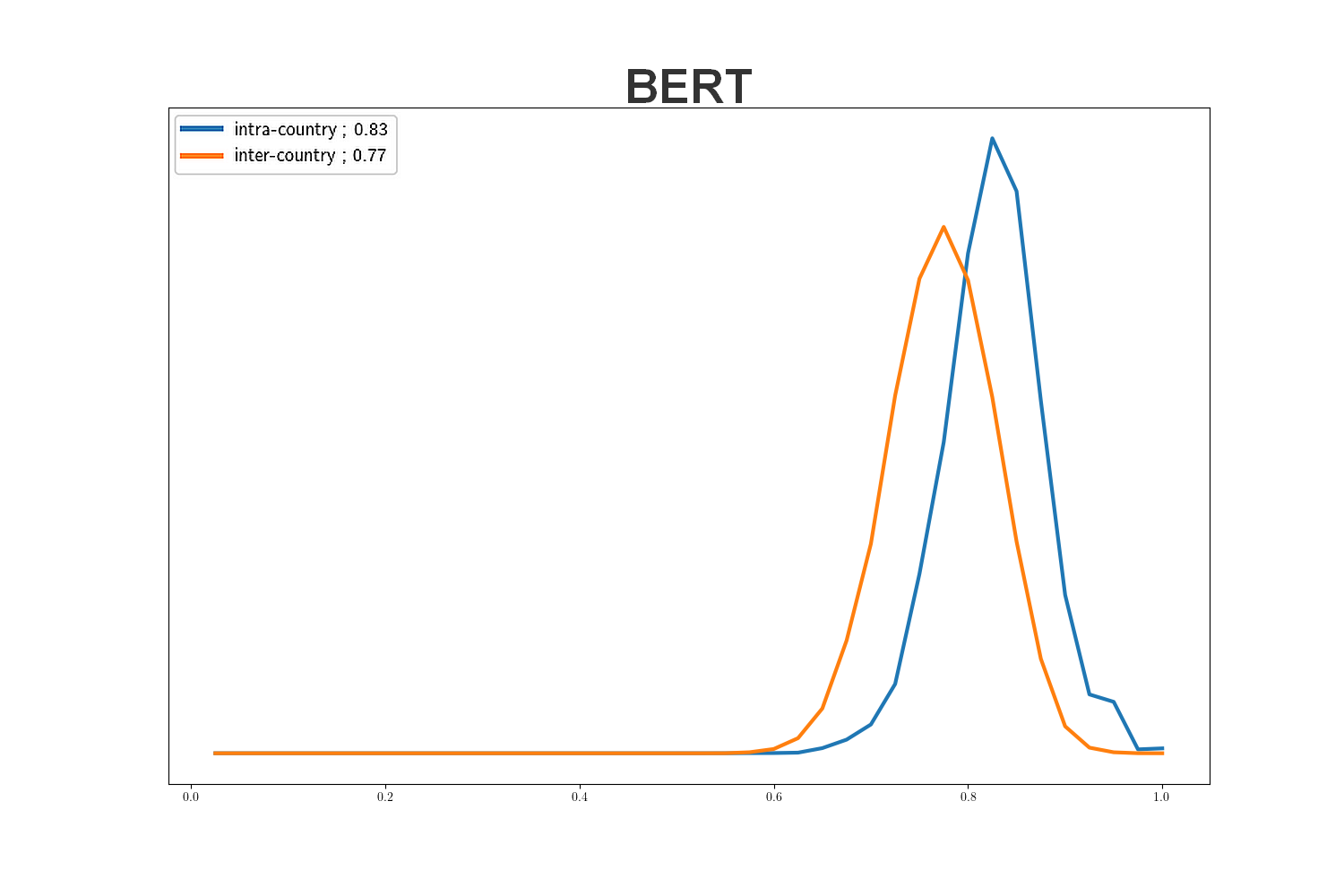}\includegraphics[width=8cm]{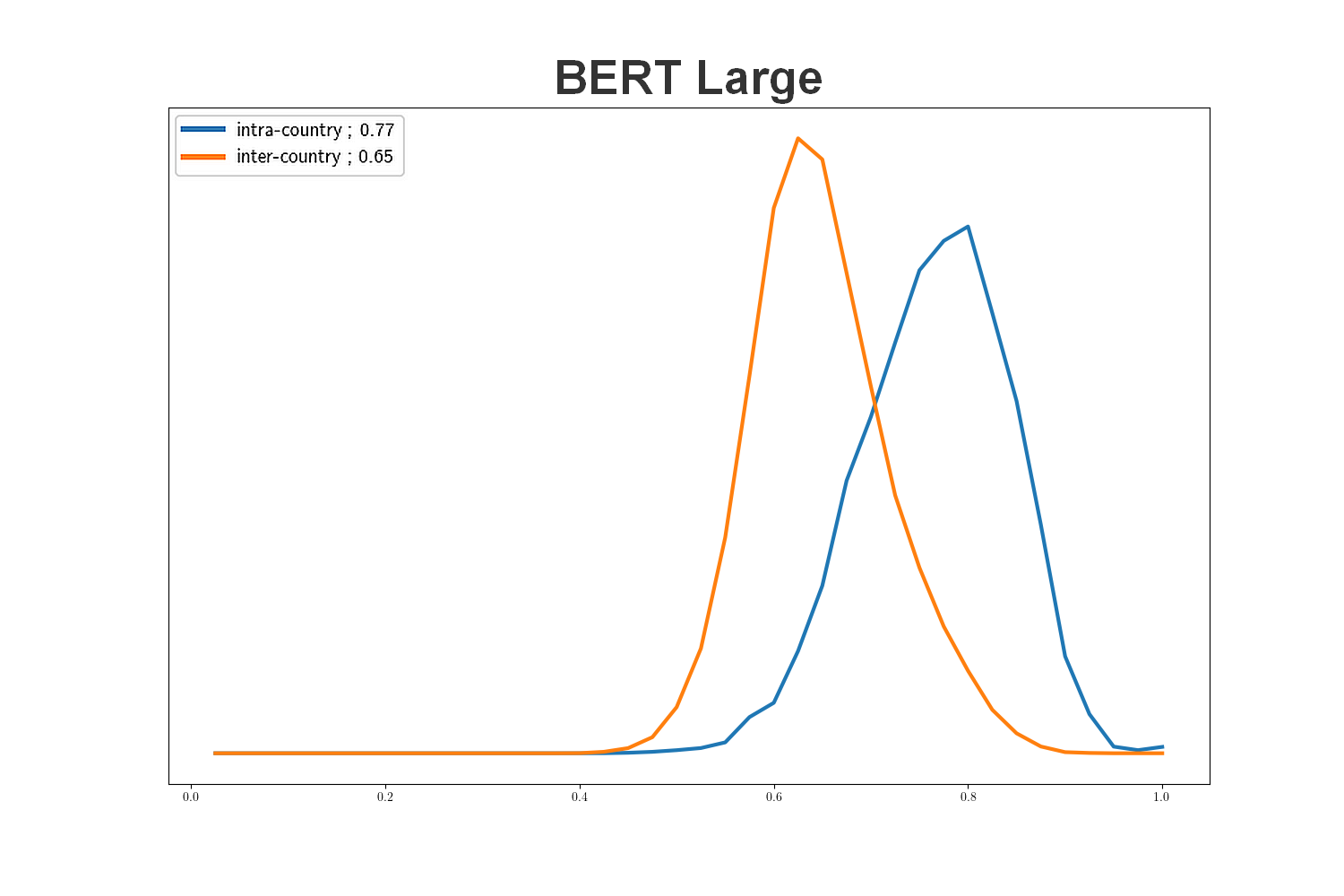}
    
    \includegraphics[width=8cm]{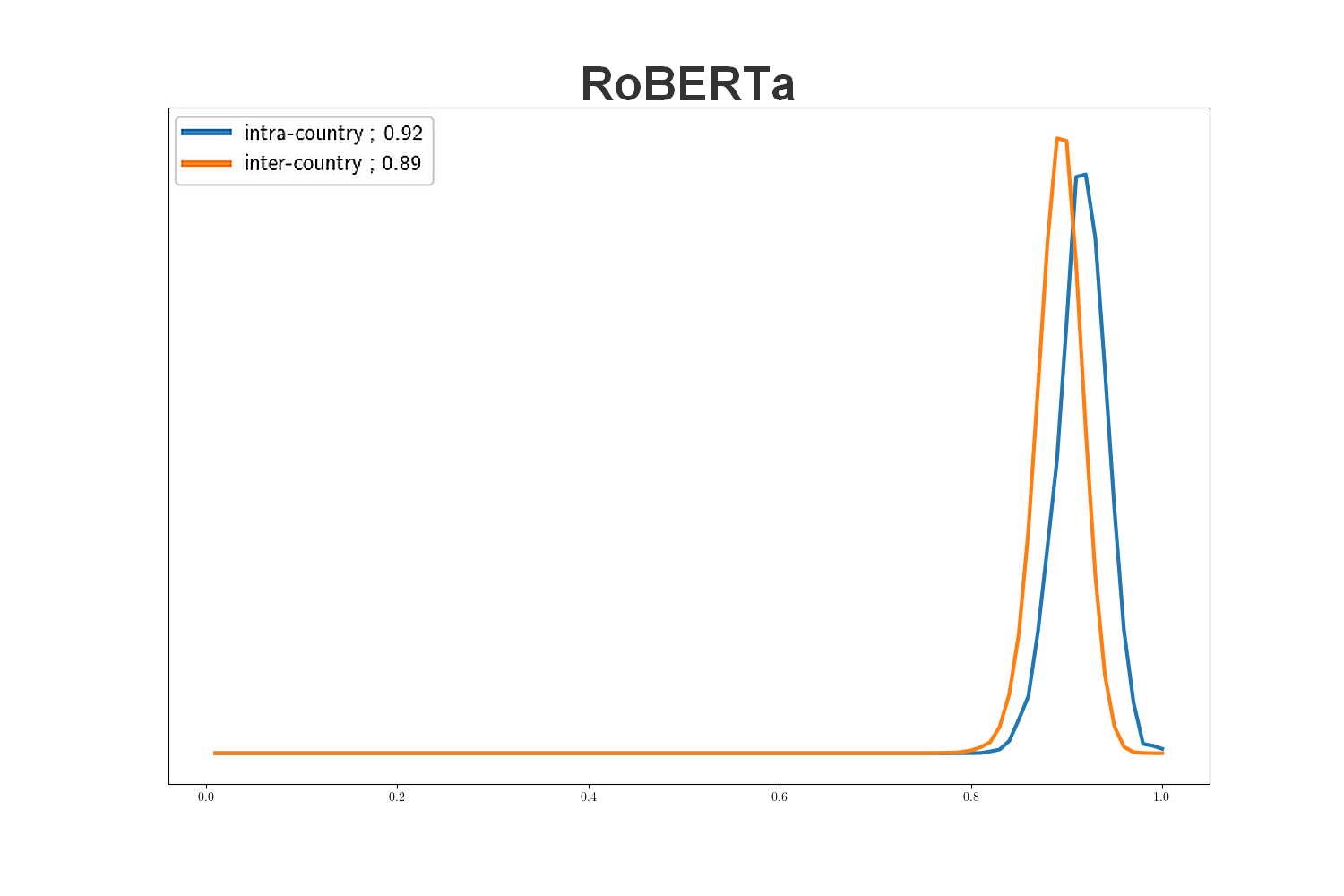}\includegraphics[width=8cm]{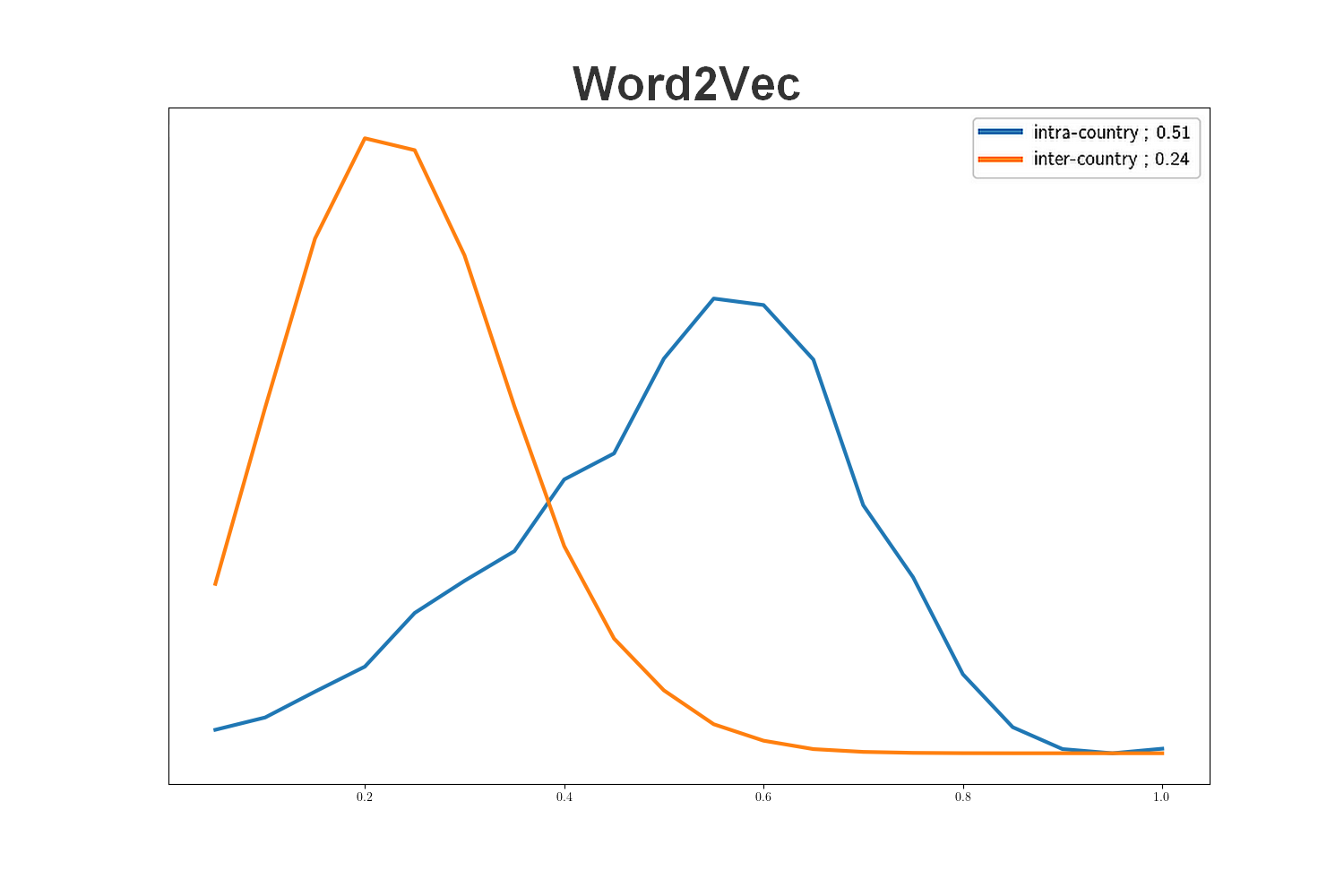}
    \caption{Representational similarity histograms for BERT, BERT \texttt{large}, RoBERTa and Word2Vec. The similarity is computed through cosine similarity. Blue is the curve of \textit{intra-country} (same country) similarity ; orange is the one for \textit{inter-countries} (different countries) similarity.}
    \label{fig:sim_hists}
\end{figure*}

\begin{table}[t]
    \centering
    \begin{tabular}{c|ccc}\hline
        Model & intra & inter & gap\\\hline\hline
        Word2Vec & 0.51 & 0.24 & 0.27\\
        BERT & 0.83 & 0.77 & 0.06\\
        BERT-L & 0.77 & 0.65 & 0.12\\
        GPT-2 & 0.90 & 0.89 & 0.01\\
        RoBERTa & 0.92 & 0.89 & 0.03\\
        RoBERTa-L & 0.94 & 0.93 & 0.01\\\hline
    \end{tabular}
    \caption{Average representational similarities between cities in the same country (\texttt{intra}), in different countries (\texttt{inter}) and the gap between them.}
    \label{tab:simil}
\end{table}

In Task 2, the gap between the two regressors is more noticeable {compared to Task 1}: MLP exhibits much higher error reductions between the original task and the random control, and generally allow for lower error numbers. We remark that our Lasso model is particularly ineffective for country population prediction, leading only to low PER and high errors. On this task and dataset, GPT-2 random control error is high, leading to an outlying PER for this LM, even if its probing-task error is still the biggest. The scheme {observed in Task 1} between \texttt{base} and \texttt{large} version of BERT and RoBERTa seems to hold, with only an exception for countries with MLP. This overall indicates that a one-dimensional attribute is easier to learn than a two-dimensional space, giving more expressive models an upper hand.

Table \ref{tab:border_probe} shows the probe's accuracy and selectivity on Task 3, predicting country borders. This is a local relation, included for comparison. Probe accuracy scores exhibit little variance, and control accuracies are almost identical, close to \textbf{0.5}, which is expected for random binary classification. All language models allow exhibit probe accuracies above $80\%$. This indicates the insufficiency of probing only for local relations.

The results of the similarity analysis on city names are displayed in Table \ref{tab:simil} and in Figure \ref{fig:sim_hists}. It is to be observed that representations are highly similar in contextualized LMs, especially compared to those from Word2Vec, which is in accordance with \citet{Ethayarajh2019} results about anisotropy in higher layers of LMs. On the other hand, we observe that representations of cities in the same country are always more similar than cities in different countries, even if the gap is small for some models. Finally, we can note that no conclusion can be made about the effect of increasing the LM's size in this analysis: the gap is bigger in large version of BERT than in the base version, but this relation does not hold for RoBERTa.

\section{Discussion}

Tasks 1 to 3 suggest that the amount of geographic knowledge learned by the language models is limited. On the one hand, trained predictors for both GPS coordinates and population are of poor quality, making average errors of thousands of kilometers or hundreds of millions of people. On the other hand, the probes are doing significantly better on these tasks than during random control, and information about country borders have been shown by results of Task 3 to be embedded in pre-trained language models' representations. This suggests that approximate geographic positions and their neighborhood are learned during training by these models. Similarity analysis also indicates that pre-trained models have learned to associate representations of cities in the same country. 

In the end, language models, whether they are static or contextualized, seem to be able to extract limited geographical knowledge from higher-order co-occurrence statistics. The size of the language models, as well as the amount of data they were trained on, seem to be important for how well they encode this information. Increasing the model's size leads to best performances and error reduction scores, indicating a better learning of geographic knowledge. Overall, context-sensitive language models do not show significantly better performances than Word2Vec. Finally, our experiments demonstrate the importance of probing for global, multi-dimensional relations that require near-isomorphism and are not easily saturated. 

In the future, it would be of interest to test models finetuned on corpora which contain a wealth of geographic information (e.g. a corpus of atlases) in order to evaluate a) whether representational alignment to real-world geography can be affected by seeing factual expressions such as \textit{``Italy, France, and Switzerland share borders."} and b) how exactly representations of entities (e.g. \textit{France}) alter when the model sees  factual statements as the one above. 

\section*{Acknowledgements}

We thank the anonymous reviewers. Mostafa Abdou was funded by a Google Focused Research Award. We used data created by MaxMind, available from
\url{http://www.maxmind.com/}. 

\bibliography{anthology,custom}
\bibliographystyle{acl_natbib}

\appendix

\section{Appendices}

\subsection{Task 1 performances}
\label{sec:task1perf}

On the following tables, \textit{Prb.} stands for ``probing task'' and \textit{Ctl.} for ``control task''.

\begin{table}[h]
    \centering
    \begin{tabular}{c|cc|cc}\hline
        \multirow{2}{*}{Model} & \multicolumn{2}{c|}{MLP} & \multicolumn{2}{c}{Lasso}\\
         & Prb. & Ctl. & Prb. & Ctl.\\\hline\hline
        Word2Vec & \textbf{2612} & 7825 & 3447 & 6870\\
        BERT & 4195 & 8057 & 3780 & 6920\\
        BERT-L & 3315 & 7997 & \textbf{3077} &6911 \\
        GPT-2 & 4613 & 8011 & 4498 & 7070\\
        RoBERTa & 4278 & 8007 & 4148 & 6894\\
        RoBERTa-L & 3876 & 8029 & 3686 & 6903\\\hline
    \end{tabular}
    \caption*{Mean error distances in kilometers for GPS prediction of \textbf{cities}}
    \label{tab:task1city}
\end{table}

\begin{table}[h]
    \centering
    \begin{tabular}{c|cc|cc}\hline
        \multirow{2}{*}{Model} & \multicolumn{2}{c|}{MLP} & \multicolumn{2}{c}{Lasso}\\
         & Prb. & Ctrl. & Prb. & Ctrl.\\\hline\hline
        Word2Vec & 3738 & 8695 & \textbf{4379} & 7234\\
        BERT & 4950 & 8598 & 4944 & 8152\\
        BERT-L & \textbf{3603} & 8578 & 4488 & 8394\\
        GPT-2 & 5111 & 8840 & 5658 & 8684\\
        RoBERTa & 5522 & 9275 & 5036 & 6091\\
        RoBERTa-L & 4764 & 8960 & 4433 & 8125\\\hline
    \end{tabular}
    \caption*{Mean error distances in kilometers for GPS prediction of \textbf{countries}}
    \label{tab:task1country}
\end{table}

\subsection{Task 2 performances on countries}
\label{sec:task2perf}

Reported results are \textit{mean squared errors}. Therefore, mean absolute errors are at most of the magnitude of few hundreds (of millions) of people.

\begin{table}[h]
    \centering
    \begin{tabular}{c|cc|cc}\hline
        \multirow{2}{*}{Model} & \multicolumn{2}{c|}{MLP} & \multicolumn{2}{c}{Lasso}\\
         & Prb. & Ctrl. & Prb. & Ctrl.\\\hline\hline
        Word2Vec & 12142 & 31815 & 22112 & 17810\\
        BERT & 16382 & 39952 & 26583 & 30063\\
        BERT-L & 17166 & 46365 & 22559 & 31174\\
        GPT-2 & 15264 & 3566 & 32130 & 51171\\
        RoBERTa & 15266 & 32338 & 26375 & 28592\\
        RoBERTa-L & 16390 & 33112 & 22927 & 25923\\\hline
    \end{tabular}
    \caption*{Mean squared error distances (in millions of people) for prediction of country's population.}
    \label{tab:task2country}
\end{table}

\end{document}